\documentclass[letterpaper, 10 pt, conference]{ieeeconf}  

\usepackage{hyperref}
\usepackage{caption}
\usepackage{amsmath}
\usepackage{amssymb}

\IEEEoverridecommandlockouts                              

\usepackage{graphicx} 

\title{\LARGE \bf
Towards Human-Centered Construction Robotics: A Reinforcement Learning-Driven Companion Robot for Contextually Assisting Carpentry Workers
}
\author{Yuning Wu$^{1}$, Jiaying Wei$^{1}$, Jean Oh$^{2}$, and Daniel Cardoso Llach$^{1}$%
\thanks{$^{1}$Computational Design Laboratory, Carnegie Mellon University, 5000 Forbes Ave, Pittsburgh, PA, USA.
        {\scriptsize\tt {\{yuningw, jwei3, dcardoso\}@andrew.cmu.edu}}}%
\thanks{$^{2}$Robotics Institute, Carnegie Mellon University, 5000 Forbes Ave, Pittsburgh, PA, USA. {\scriptsize\tt jeanoh@cmu.edu}}
}

\begin{document}

\maketitle
\thispagestyle{empty}
\pagestyle{empty}

\begin{abstract}

In the dynamic construction industry, traditional robotic integration has primarily focused on automating specific tasks, often overlooking the complexity and variability of human aspects in construction workflows. This paper introduces a human-centered approach with a ``work companion rover" designed to assist construction workers within their existing practices, aiming to enhance safety and workflow fluency while respecting construction labor's skilled nature. We conduct an in-depth study on deploying a robotic system in carpentry formwork, showcasing a prototype that emphasizes mobility, safety, and comfortable worker-robot collaboration in dynamic environments through a contextual Reinforcement Learning (RL)-driven modular framework. Our research advances robotic applications in construction, advocating for collaborative models where adaptive robots support rather than replace humans, underscoring the potential for an interactive and collaborative human-robot workforce.

\end{abstract}

\section{Introduction}

Construction remains the world's most labor-intensive industry \cite{shehataImprovingConstructionLabor2011}, characterized by its highly manual nature and bespoke building needs. Despite advancements such as prefabrication \cite{fagbenroInfluencePrefabricatedConstruction2023} and on-site 3D printing \cite{wuCriticalReviewUse2016}, the majority of construction tasks still rely on the heavy labor of skilled workers, operating in environments that are often complex, dynamic and cluttered. These workers are not only subjected to intense physical strains but also required to perform versatile on-site improvisations. Over the past decades, a significant body of robotic research \cite{gharbiaRoboticTechnologiesOnsite2020} has aimed at creating task-specific systems designed to either entirely or partially automate work trades in construction, such as floor leveling \cite{liuBriefReviewRobotic2018}, spray painting \cite{asadiPictobotCooperativePainting2018}, and bricklaying \cite{fangArchitecturalFrameworkDistributed2020}. However, these robot-centric solutions frequently encounter difficulties in actual on-site integration. This is largely due to the varied and flexible nature of construction work, which typically involves a complex array of diverse tasks. These automated systems often lack the necessary adaptability to handle every nuance and provide an in-situ solution comparable to human workers. As a result, they do not always effectively translate into practical, real-world applications. Furthermore, the practicalities of operating, maintaining, and supplying these systems on an actual construction site, as opposed to an idealized laboratory environment, remain challenging.

\begin{figure}[!htbp]
  \centering
  \includegraphics[width=\linewidth]{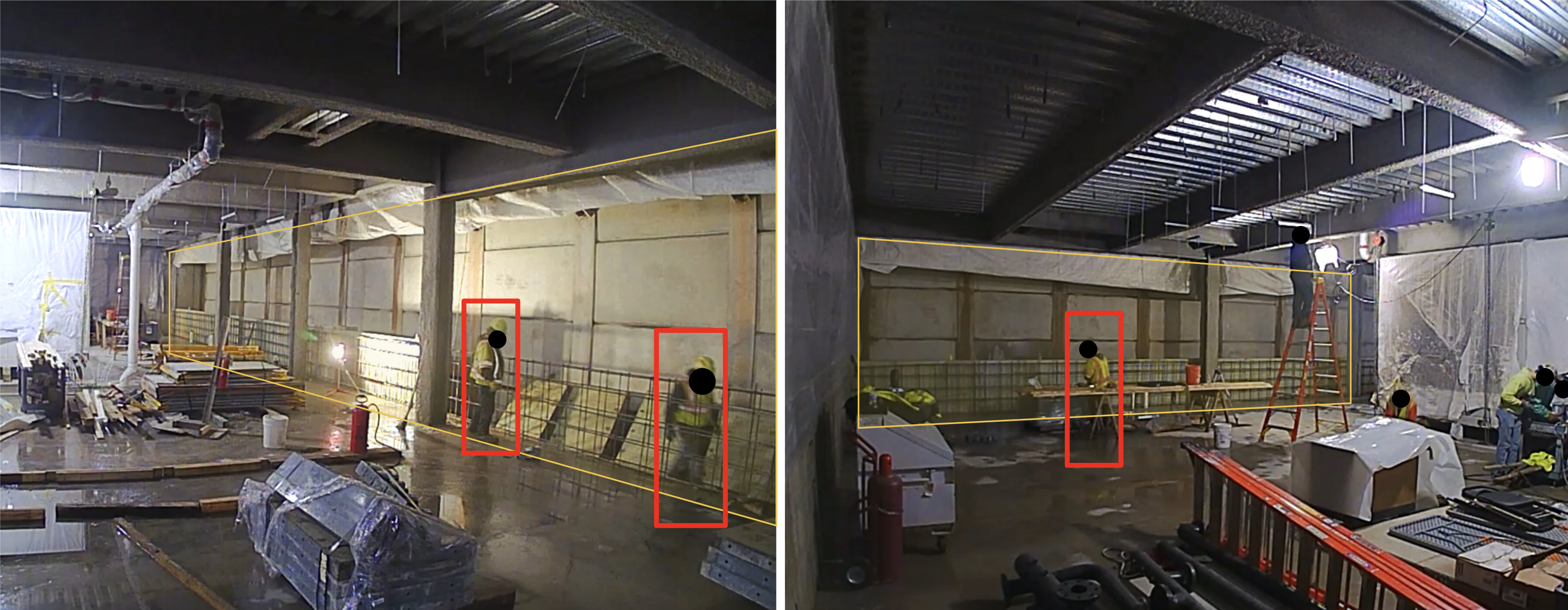}
  \caption{Site observation captures of the carpentry formwork activitity and construction site condition}
  \vspace*{-5mm}
  \label{fig:site-obs}
\end{figure}

Drawing from the lessons of past endeavors in construction robotics \cite{bockFutureConstructionAutomation2015}, our research adopts a human-centric approach, informed by the latest developments in deep reinforcement learning (RL) \cite{suttonReinforcementLearningIntroduction2018}. This approach diverges from the traditional aim of complete automation of manual construction processes. Instead, our study introduces an ecological integration of mobile robots into existing manual workflows, positioning them in assistive and supportive roles. By undertaking tasks that are physically demanding yet seemingly minor, such robots can emerge as mobile work companions, allowing human workers to concentrate on the more skilled and critical aspects of their work. This hybrid, transitive model of robotically supported work collaboration seeks to address enduring construction challenges, including reducing physical strain, mitigating workplace injuries, and enhancing workflow fluency.

Based on in-depth observations at an actual construction site (Fig. \ref{fig:site-obs}), we focused on carpentry formwork—a labor-intensive and prevalent construction activity—as a prototypical scenario to explore the envisioned robotic support. Informed by a qualitative study of the social and material specifics of this work scenario, we developed a prototype, termed "work companion rover", aimed at providing tangible support to a duo of carpentry formwork installation workers. This rover's support functions encompass autonomous delivery of tools and materials, weight-bearing capabilities, and companionship during work tasks. Following its development, the prototype underwent both qualitative and quantitative evaluations to assess its support capabilities, conducted both in lab settings and on an actual construction site with workers.

This paper's contributions are threefold: (1) introducing a human-centered ``work companion rover" prototype, specifically designed to closely support carpentry workers in their existing, labor-intensive tasks, (2) developing a lightweight, modular, and expandable framework driven by RL-based social navigation methods that can foster safe and comfortable navigation of mobile robots in real-world construction environments, broadly construed, and (3) showcasing a practical and efficient pipeline for contextually aligning and improving generically pretrained RL models with context-specific features. At a broader level, we aim to illustrate through these contributions a feasible, alternative pathway for integrating autonomous robots into construction and labor-intensive work. This approach strategically values human skill and expertise, while concurrently harnessing AI and robotics to improve work safety and workflow fluency.

\section{Background and Related Works}

\subsection{Assistive robots in various domains}

In the realm of robotics, decades of research and development have been dedicated to assistive robots and human-robot teaming \cite{mingyuemaHumanRobotTeamingConcepts2018} for various domains. Notable fields include healthcare \cite{calderitaTHERAPISTAutonomousSocially2014, fasolaSociallyAssistiveRobot2013}, public guiding and touring \cite{iioHumanLikeGuideRobot2020}, disaster response and rescue \cite{kruijffExperienceSystemDesign2014}, space exploration \cite{hopkoHumanFactorsConsiderations2022}, etc. In these contexts, such technologies have played a critical role in enhancing service quality and user experience. However, their application in labor-intensive work sectors, particularly in on-site construction, remains relatively sparse. This oversight is significant given that construction is not only the largest industry worldwide but also one that is heavily manual, and often operates within physically demanding and complex indoor environments.

Our recent investigation at an active construction site revealed significant potential for robotic assistance within established workflows. Implementing such assistance on construction sites may reduce manual traversals, diminish physical fatigue in workers through sensitive support, and improve site safety by alleviating ongoing concerns about workplace injuries. To elucidate this viewpoint, we carefully observed eleven trades on-site during the ``rough carpentry stage"—a construction phase characterized by the completion of concrete floors and major structural elements, concurrent with intensive interior manual tasks and installations. We focused particularly on carpentry formwork, a labor-intensive process typically conducted by a two-person team, involving the manual installation of formwork panels to create external molds for concrete walls. Our observations revealed that workers, in addition to their primary installation work, spend considerable effort in transporting tools, materials, and hardware in small batches to their work zones, either from a distant central workbench or a previous work area. Within a single hour, we noted upwards of 20 such traversals. This frequent activity in the current workflow presents a major yet overlooked physical burden for workers. It is an opportune area for the introduction of a "work companion rover", capable of autonomously delivering tools and hardware upon simple interactions, as well as comfortably accompanying workers unobtrusively by carrying heavy loads of frequently used objects nearby.

\subsection{Robot navigation and mobility on construction site}

Existing research in the field of robot navigation has shown considerable progress over the past decades \cite{pandeyMobileRobotNavigation2017}. However, the majority of the research, with a few notable exceptions, is based on contexts that differ from ours in terms of robotic functionality and application. For example, Logistic robots, which are increasingly common in factories and warehouses, are designed for efficient goal-directed deliveries in environments that are far more orderly and organized than the chaotic and confined spaces of indoor construction sites. A substantial amount of research in social robot navigation focuses on enabling robots to navigate among densely populated areas with social compliance \cite{mavrogiannisCoreChallengesSocial2021}, with applications such as robot delivery on urban streets \cite{chenAdoptionSelfdrivingDelivery2021} and university campuses \cite{williamssonBusinessModelDesign2022}. In the construction domain, recent advancements in quadruped robots \cite{wetzelUseBostonDynamics2022} demonstrate their potential to autonomously inspect, monitor, and document site progress, even across multiple floors. However, these capabilities are not tailored to directly provide human-centered, integrated support to heavy manual laborers within their current workflows.

Acknowledging these fundamental differences in context and functions, we identified two primary challenges specific to our research. Firstly, the robot must be capable of safely navigating through physically unstructured and complex construction environments, especially the floor area relevant to carpentry formwork. Secondly, to effectively provide support to carpentry workers, the robot should be able to navigate comfortably and closely around workers, taking into account their specific work activities and features. Tackling these challenges demands a nuanced understanding of both the physical and social aspects of robot navigation within the unique context of on-site construction work.

\subsection{Reinforcement learning for social robot navigation}

Recent advances in social robot navigation \cite{singamaneniSurveySociallyAware2024} have demonstrated progress in addressing the social aspect of the aforementioned challenges. Notably, reinforcement learning is emerging as a promising approach for handling and adapting to the complexities associated with the dynamic movements of neighboring agents. Various deep neural network (DNN) based algorithms have been developed to tackle diverse social dynamics and challenges. For example, initial studies \cite{everettMotionPlanningDynamic2018} employed Long Short-Term Memory (LSTM) networks to capture latent state encodings of neighboring agents' movements. Subsequent research \cite{chenCrowdrobotInteractionCrowdaware2019} has expanded this scope to include human-human interactions in addition to human-robot interactions, thereby enriching the learning process. More recent studies \cite{liuDecentralizedStructuralRNNRobot2021} have employed decentralized structural Recurrent Neural Networks (RNNs) to explicitly model the spatial and temporal relationships within crowds. Additionally, some research \cite{stolerT2FPVDatasetMethod2023} has concentrated on the first-person perspective of social robot navigation, notably benchmarking the impact of this limited viewpoint on navigation performance. Despite these algorithmic advancements, there remain notable gaps in applying these methods effectively to our specific research context and functionality. Most notably, this body of algorithmic work typically assumes an open-space layout, devoid of complex physical obstacles, and therefore would require significant additional effort in forming a comprehensive solution. Moreover, these works are mostly situated in dense crowd-based settings, where agents are mostly modeled as goal-reaching pedestrians.

The fundamental challenge of our research lies in how to meaningfully leverage the existing efforts in RL-based social robot navigation research while contextualizing them with the intricacies of construction work and site. This requires not only designing a framework to account for the complicated site surroundings but also ensuring that the social dynamics of human-robot interactions remain effective in more challenging work settings. Importantly, carpentry workers are not pedestrians and have drastically different activity patterns than simple walking. The integration of these two aspects—navigating a physically complex environment and contextually adapting to worker-specific interactions and comforts—will be critical to the success of our research.

\section{Problem Statement}

\subsection{Robotic support scenarios and functions}

\begin{figure}[!htbp]
\centering
\includegraphics[width=\linewidth]{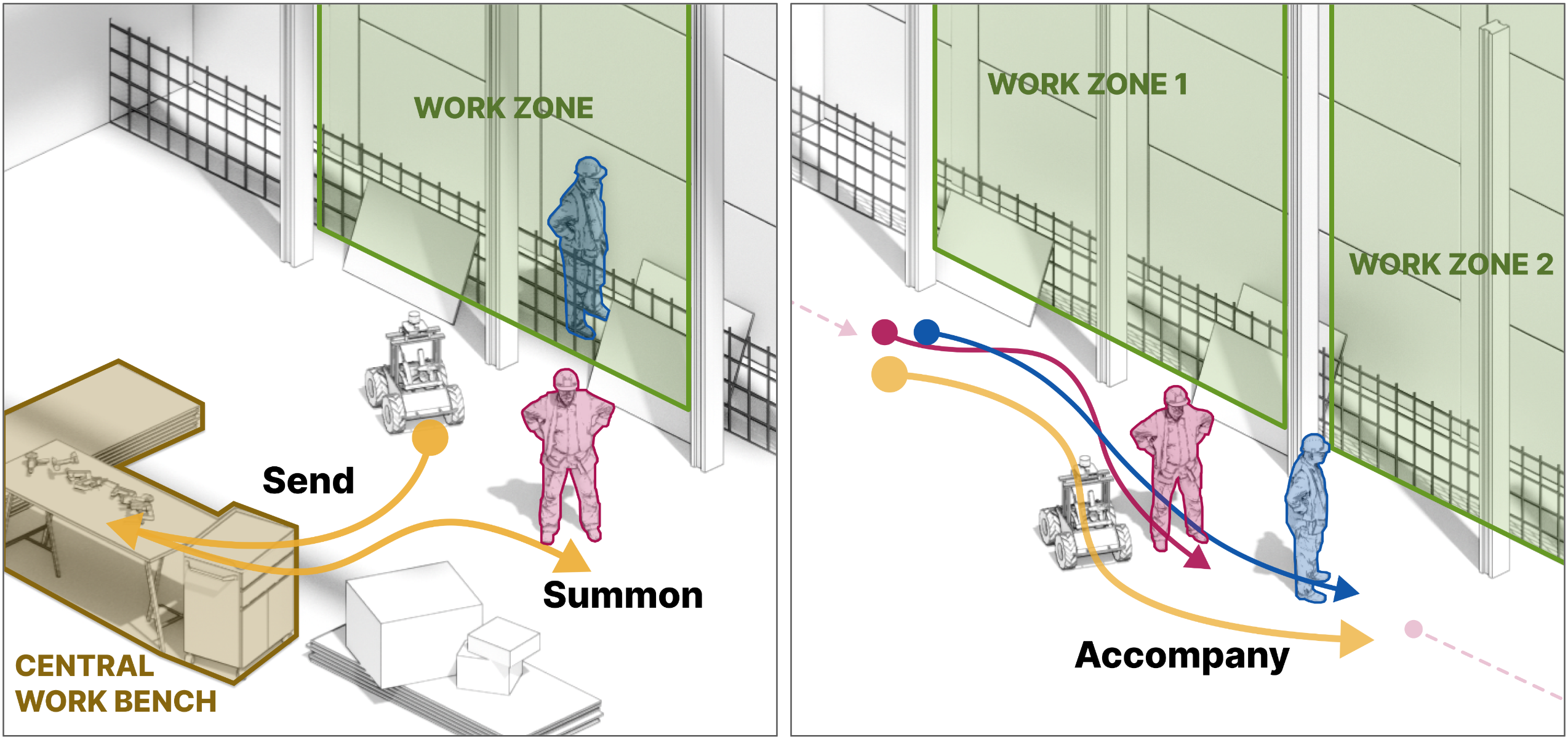}
\caption{The robotic support scenarios and functions}
\vspace*{-3mm}
\label{fig:site-scenario}
\end{figure}

To concretely realize and validate the envisioned robotic support, we identified three key robot functions within two primary support scenarios. The first scenario involves tool/hardware delivery between a distant workbench and the workers' current work zone (left in Fig. \ref{fig:site-scenario}), while the second scenario encompasses close-distance load-bearing and accompanying as workers migrate between adjacent work zones (right in Fig. \ref{fig:site-scenario}). The three fundamental robot functions essential to these scenarios are \textit{send}, \textit{summon}, and \textit{accompany}. The \textit{send} function allows workers to dispatch the robot to the central workbench, approximately 20m away, for tools and materials. The \textit{summon} function enables workers to remotely call back the robot after tools and materials have been loaded. The \textit{accompany} function aids workers as they transition to the next work zone by preventing the need for workers to manually lift and transport tools and materials. The robot closely accompanies the workers, carrying the loads in its container for easy access. The robot is expected to perform these functions with a focus on navigation comfort, not to cause interruption or obstruction to the workers' ongoing activities. Moreover, these functions are intended to be easily operable by workers through simple, intuitive commands, such as a single key press on a controller.

\subsection{MDP formulation}

The essence of the above robot scenarios and functions can be modeled as a 2D navigation task where a robot moves towards an anticipated goal position while encountering and interacting with $n$ workers in a cluttered physical layout $L$. Following a similar formulation in \cite{chenCrowdrobotInteractionCrowdaware2019, liuDecentralizedStructuralRNNRobot2021}, we model the worker-robot interaction scenarios as a Markov Decision Process (MDP) of $\langle \mathcal{S}, \mathcal{A}, \mathcal{P}, \mathcal{R}, \gamma, \mathcal{S}_0\rangle$. The observable state for both worker and robot includes position $\mathbf{p}=[p_x, p_y]$, $\mathbf{v}=[v_x, v_y]$, and safety radius $r$. For the robot, we assume it is aware of its preferred speed $v_{pref}$ and goal position $\mathbf{p}_g=[p_{gx}, p_{gy}]$, and that $\mathbf{v}_t = \mathbf{a}_t$. At time $t$, the joint state for the robot encompasses both its own full state $\mathbf{s}_t^{\text{full}}$ and the observable states of all neighboring workers $\mathbf{w}_t^{\text{obs}}$, $\mathbf{s}^{\text{joint}}_t = [\mathbf{s}_t^{\text{full}}, \mathbf{w}_t^{\text{obs}}]$. The MDP can be solved through reinforcement learning to maximize the expected return, $R_t=\mathbb{E}\left[\Sigma_{k=t}^T \gamma^{k-t}r_k\right]$. Our goal is to optimize this objective while taking contextual information of cluttered layout $L$ and workers' unique activity patterns into consideration.

\section{Approaches}

\begin{figure}[!htbp]
  \centering
  \includegraphics[width=\linewidth]{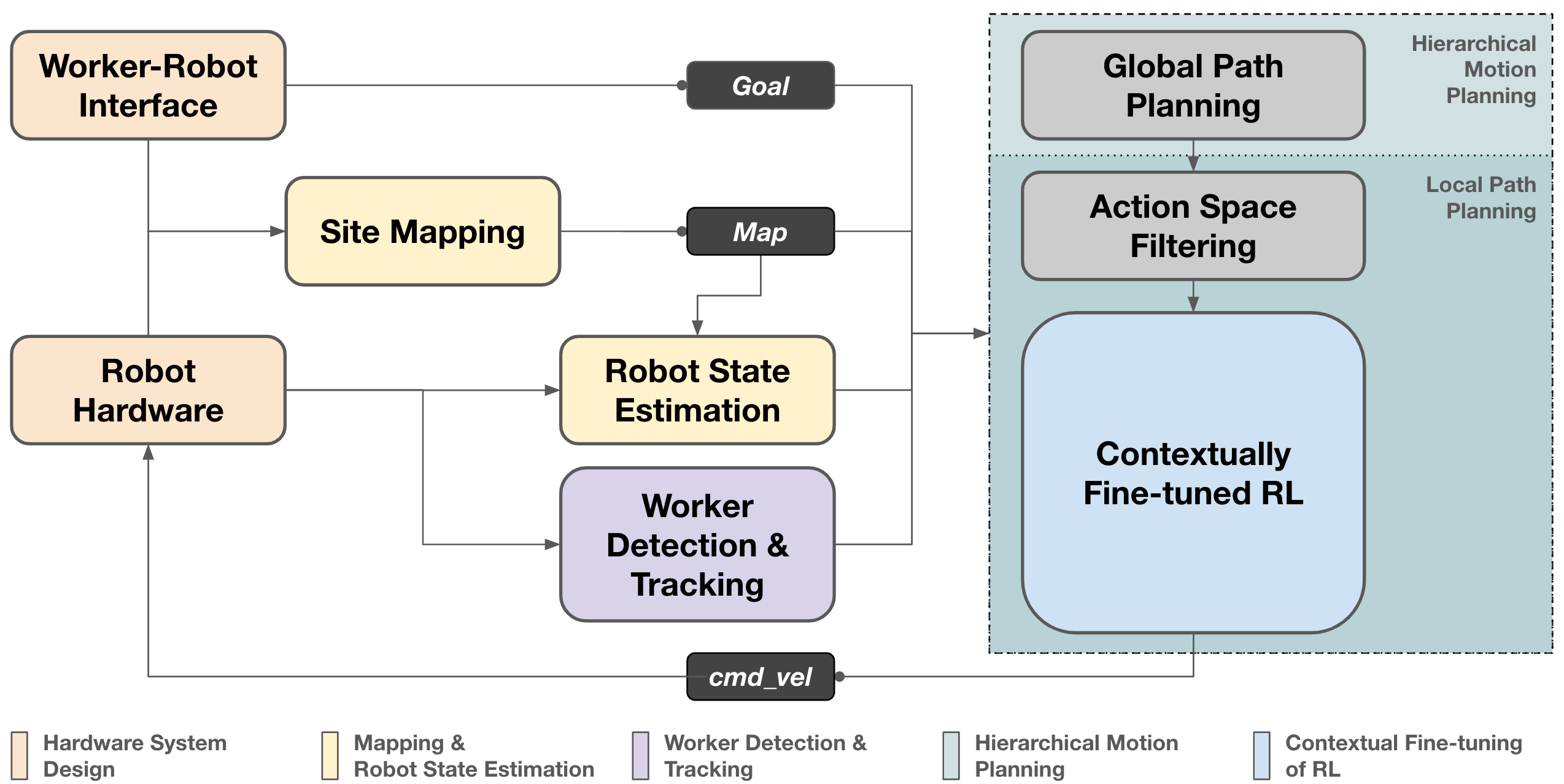}
  \caption{The modular framework design}
  \vspace*{-3mm}
  \label{fig:framework}
\end{figure}

In response to the described problem and challenge, our research at a high level adopts a research-by-prototyping approach. Drawing insights from site observations and worker interviews, we developed a middle-sized companion rover prototype suitable for close-range operation around workers from a generic UGV base, and used the prototype for real-world experimentation of the envisioned support. The prototyping process commenced with the development of a modular system framework (Fig. \ref{fig:framework}), designed to be both flexible and adaptable to the specific challenges of a construction work environment. This framework is built upon the ROS navigation stack with context-specific customization, incrementally integrating modules to address (1) the complexities of the construction site, (2) the perception of carpentry workers' activities, and (3) the contextual adaptation of RL-based social navigation methods. In more detail, the framework comprises five key modules: (1) hardware system design, (2) site mapping and robot state estimation, (3) worker detection and tracking, (4) hierarchical motion planning, and (5) contextual fine-tuning of RL. Each module will be detailed below with a focus on addressing context- and domain-specific challenges.

\subsection{Hardware system design}
\begin{figure}[!htbp]
  \centering
  \includegraphics[width=\linewidth]{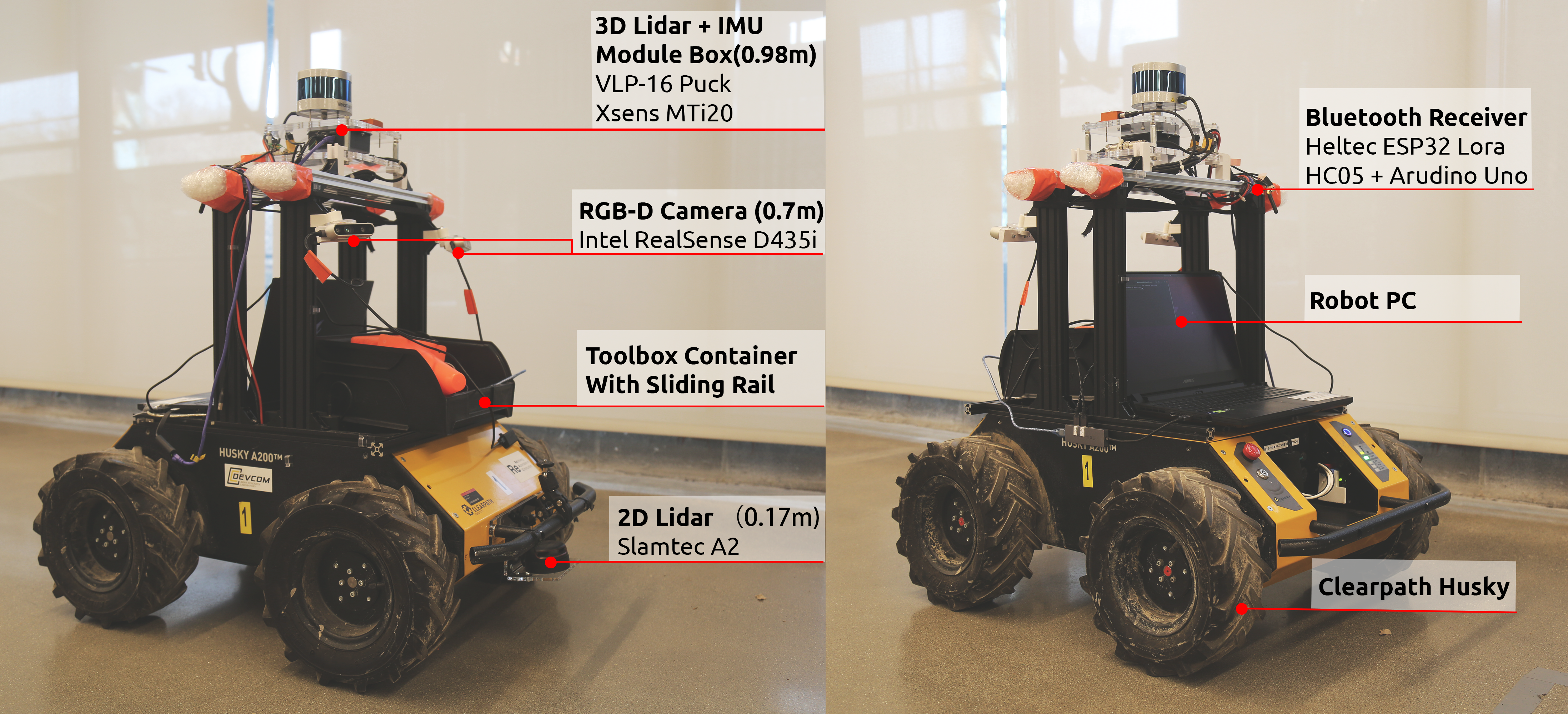}
  \caption{Hardware design of the robot}
  \vspace*{-3mm}
  \label{fig:robot-hardware}
\end{figure}

The challenging conditions of a construction site, coupled with the demands of tool and material carriage, have necessitated critical considerations in the development of our robot's hardware (Fig. \ref{fig:robot-hardware}). During the carpentry formwork stage, the construction floor, while mostly flat, is often cluttered with dust, puddles, cables, and various low-lying obstacles such as hand tools and material piles. To navigate fluently on such rough surfaces, the robot chassis needs to be both robust and agile. With this preliminary in mind, we evaluated three different robot bases: a Fetch robot, a custom-built omnidirectional robot, and a Clearpath Husky robot. Our trials revealed that the Fetch robot and the omnidirectional robot, both limited by smaller wheel sizes, faced significant challenges when maneuvering over cables, potholes, and coarse sawdust. In contrast, the Clearpath Husky robot, which was originally designed for rugged outdoor terrains, emerged as the superior choice for our application. The Husky robot is equipped with large, thick, and durable wheels and differential control motors, enhancing its ability to navigate over uneven surfaces and obstacles. Additionally, its low center of gravity contributes to stability, which is critical when bearing and transporting tools and materials. The inclusion of two T-slot rails on the robot also allows for the installation of additional structures or equipment, providing the flexibility of customization needed for our context.

The sensor package atop the robot base is a multi-modal assembly designed for navigating the geometric complexities of a construction site and perceiving worker activities. A 3D LiDAR (Velodyne VLP-16) is positioned on top of the robot frame for maximal point cloud data collection, coupled with an Xsens IMU. A 2D LiDAR (RPLiDAR A2), placed at the robot's front bottom, compensates for the 3D LiDAR's blind zone, ensuring the detection of objects as low as 8cm (e.g., hand tools, materials, and large debris) in the robot's path. Additionally, two Intel RealSense D435i RGB-D cameras are installed on the sides for effective worker detection and tracking.

The robot integrates an onboard computer with an NVIDIA RTX 3070 Ti GPU for real-time inference of multiple DNNs. It also features a 35cm x 45cm retractable container on the deck, sized based on common carpentry tools and materials. After sensor addition, the maximum payload of the robot is 35 kg on concrete surfaces, accommodating up to 2-3 bags of metal hardware and 2-3 types of hand tools like rotary drills, pliers, and hammers. Additionally, a Bluetooth keypad controller customized for carpentry workers enables the robot to operate conveniently even with work gloves, enhancing worker-robot interaction in noisy environments.

\subsection{Mapping and localization on unstructured site}

The robot's effectiveness in supportive navigation is fundamentally anchored in quality mapping and robot state estimation within the geometrically sophisticated construction work environment. Objects on the site range from very tall (e.g., ladders, steel columns, frames, material stacks, boxes) to very low (e.g., hand tools, material piles, waste clusters, randomly placed extension sockets), and are inevitably distributed in an unpredictable manner. This setting diverges significantly from more structured environments like warehouses, manufacturing factories, or open public spaces.

\begin{figure}[!htbp]
  \centering
  \includegraphics[width=\linewidth]{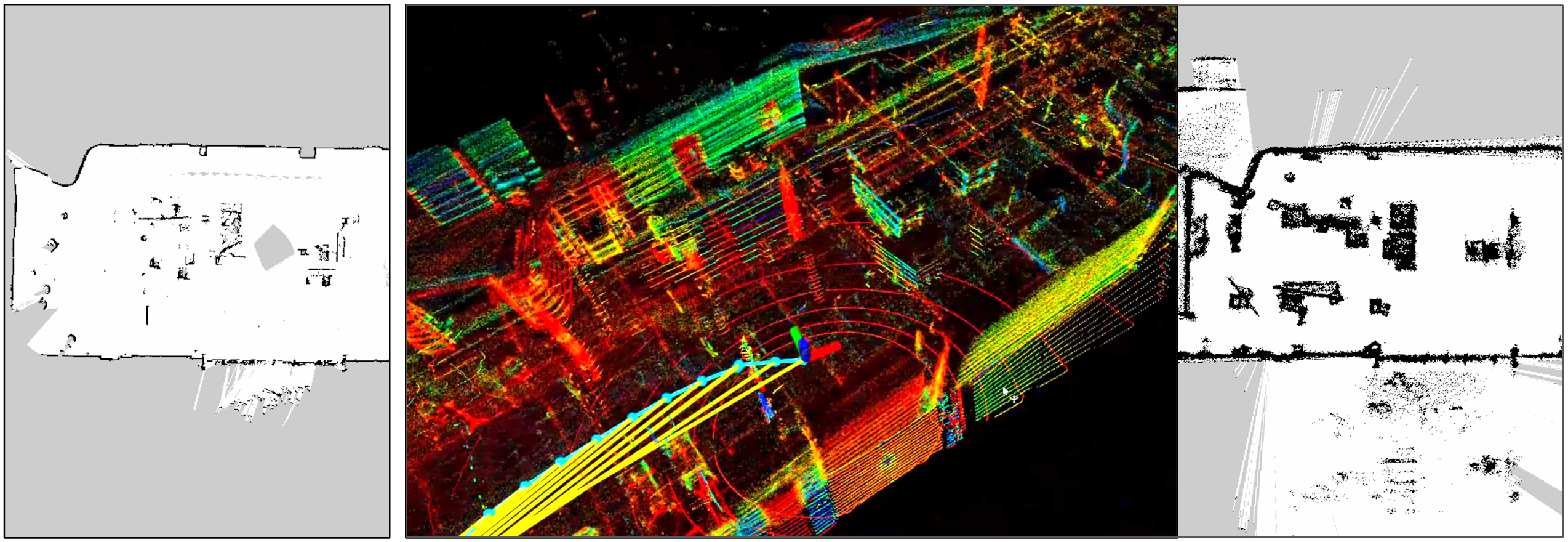}
  \caption{Grid map from \texttt{gmapping} (left) and ours (right)}
  \vspace*{-3mm}
  \label{fig:mapping}
\end{figure}

Conventional 2D mapping tools, such as \texttt{gmapping}, are insufficient for capturing the complex topography of a construction site (Fig. \ref{fig:mapping}), such as height-varying objects (e.g., open ladders, tool tripods) and randomly extruding hazards (e.g., horizontally stacked steel beams and pipes). Meanwhile, 3D voxel-based or point cloud-based representations are too computationally inefficient for downstream navigation tasks. To address this, we employ a recent LiDAR Odometry and Mapping (LOAM) \cite{zhangLOAMLidarOdometry2014, shanLIOSAMTightlycoupledLidar2020} method to generate a comprehensive 3D point cloud of the site by manually navigating the robot around the intended operational area with a joystick for 1-2 loops. The resulting point cloud is then projected and processed into a 2D grid map, effectively balancing the accuracy of 3D mapping with the computational efficiency of 2D methods, crucial for capturing environmental features in our context.

The cluttered and dynamic nature of construction sites also poses a challenge for the robot's state estimation, i.e., accurately determining its position and velocity with the processed map. Conventional scan matching methods like AMCL are prone to errors and drifts due to the non-stationary nature of construction objects. These objects, such as material piles, are temporary and change over time. To overcome this without constant map updates, we integrated multiple odometry and corrective information sources into the localization process through an Extended Kalman Filter (EKF), including IMU data, odometry from LOAM, the robot's wheel encoders, and AMCL.

\subsection{Worker detection and tracking}

To ensure comfortable navigation around carpentry workers, it is imperative that the robot promptly recognizes and tracks nearby workers' movements (i.e., position, velocity, and radius) against a complex backdrop, especially within its immediate action space. Drawing insights from \cite{everettRobotDesignedSocially2017}, we break down this task through detection and tracking.

Our qualitative physical experiments revealed that detection methods relying solely on 3D LiDAR data, although effective in contexts like autonomous driving and crowd detection, falter in our context. The failure is caused by the unstructured and unpredictable nature of construction objects. In many cases, the scan appearance of randomly stacked materials can mimic that of a person, leading to a high false positive rate. As an alternative, we found that vision-based methods offer greater reliability and more computational efficiency, particularly when a worker's face is within the camera's field of view (FoV). Consequently, we adopted YOLOv7 \cite{wangYOLOv7TrainableBagoffreebies2022} for worker detection, and utilized depth information to estimate the workers' position, velocity, and radius. To overcome a single RGB-D camera's limited FoV, we installed two cameras and amalgamated their detection results by clustering \cite{schubertDBSCANRevisitedRevisited2017}. Considering that the robot's low height can hinder the visibility of a worker's face when they are close to the robot, we tilted the RGB-D cameras upwards by 15 degrees to physically maintain a better vantage point. As a last step, the raw detection results were processed with Multi-Object Tracking (MOT) algorithms \cite{wojkeSimpleOnlineRealtime2017} to reduce noise and ensure worker tracking continuity, especially given frequent visual obstructions by site objects and structures.


\subsection{Hierarchical motion planning}

The robot's motion planning employs a hierarchical approach, integrating both classical methods and RL-based techniques for navigation in a cluttered and populated environment. In the local path planning stage, a lightweight search-based method, such as the Dynamic Window Approach (DWA) \cite{foxDynamicWindowApproach1997}, first serves as a foundational safety layer, efficiently filtering out actions that exceed the robot's dynamic constraints or risk collisions with people or objects (Eq. \ref{eq:1}). Rather than utilizing the efficiency-driven scoring mechanism typical of these classical methods, a trained RL value network is exploited. This network, optimized for social comfort and compliance, is used to sample and select the most appropriate final action from the refined action space $\mathcal{A}_f$ (Eq. \ref{eq:2}).

\begin{align}
    \mathcal{A}_f & = \mathcal{A} \cap \mathcal{A}_{\text{admissible}} \cap \mathcal{A}_{\text{dynamic}}\label{eq:1} \\
    \mathbf{a}_t & \leftarrow \text{argmax}_{a_t \in \mathcal{A}_f} R\left(\mathbf{s}_t^{\text{joint}}, \mathbf{a}_t\right) + \gamma^{v_{\text{pref}} \cdot \Delta t} V^{\text{joint}}_{t+\Delta t}\label{eq:2}
\end{align}

This hierarchical approach offers two key benefits: generalizability and efficiency. It enables practical comparisons among different classical and RL methods in real-world experiments with minimal engineering adjustments. Also, to ensure seamless integration between the two layers, we employed measures such as fine-tuning the searching sparsity of DWA and accelerating the parallel value network inference for actions.

\subsection{Contextual fine-tuning of generic RL model}

Through preliminary on-site observations, we identified distinct worker behavior attributes compared to those typically assumed in crowd-based social navigation research. (1) Worker groups are small (less than 5), operating densely in compact spaces. (2) Their movements are dictated by task requirements rather than efficiency, often resulting in unique patterns. (3) Engrossed in tasks, workers pay minimal attention to the robot, underscoring the need for enhanced safety.

\begin{figure}[!htbp]
  \centering
  \includegraphics[width=\linewidth]{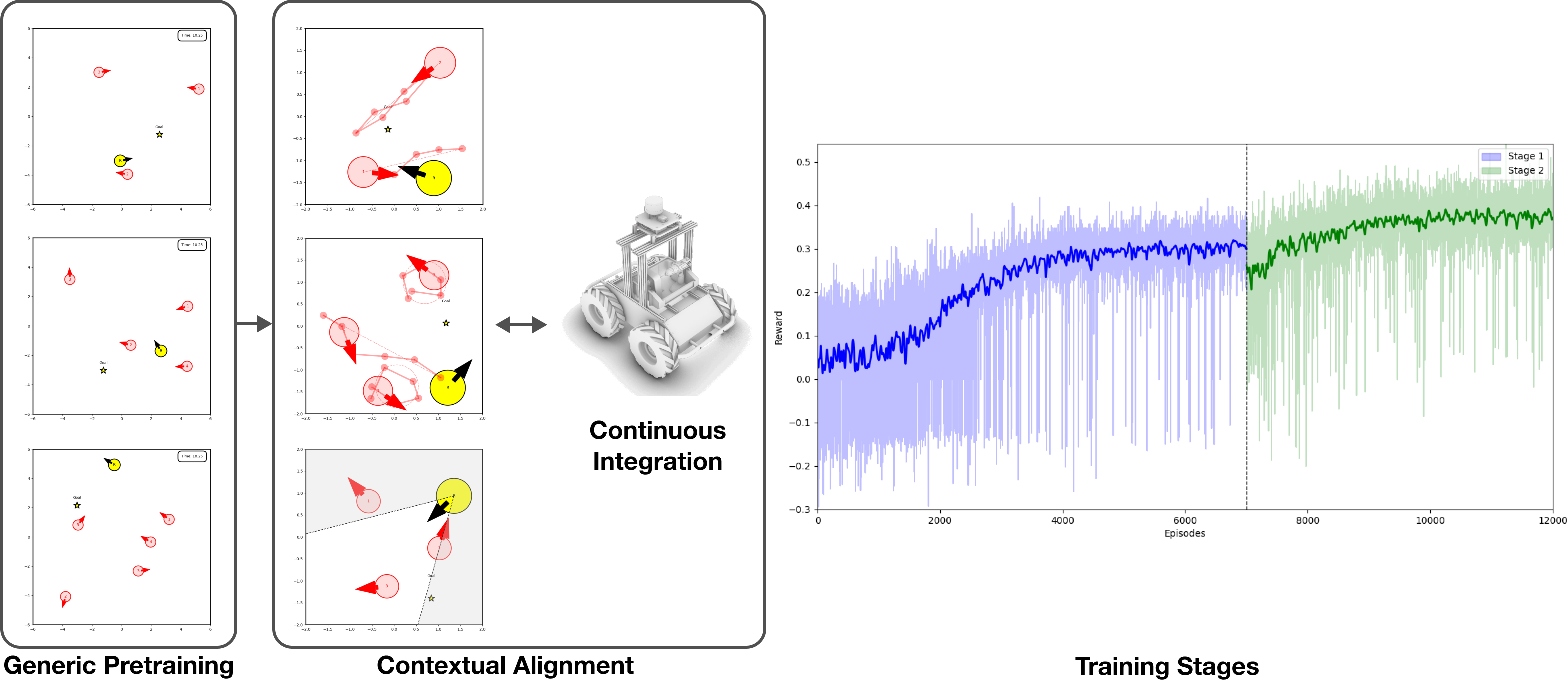}
  \caption{Contextual curriculum alignment}
  \vspace*{-3mm}
  \label{fig:rl}
\end{figure}

Acknowledging these differences, we leveraged existing RL-based social navigation research as a foundation. After reviewing and testing various algorithms on the robot, including CADRL\cite{chenSociallyAwareMotion2017}, GA3C-CADRL\cite{everettMotionPlanningDynamic2018}, LSTM-RL\cite{chenCrowdrobotInteractionCrowdaware2019}, SARL \cite{chenCrowdrobotInteractionCrowdaware2019}, etc, SARL was selected for subsequent contextual fine-tuning based on preliminary performance. Instead of starting anew, we adopted an incremental approach for model improvement and adaptation through curriculum training (Fig. \ref{fig:rl}), aiming for continuous real-world testing without the need to retrain or transfer entirely new models each time. This strategy focuses on the robot's performance stability in Sim2Real, aiming to maintain tested features while addressing observed challenges step-by-step.

To simulate the noted unique worker behaviors, we crafted and adapted context-specific experimental scenarios in the simulation environment \cite{chenCrowdrobotInteractionCrowdaware2019} (Fig. \ref{fig:rl}). Acknowledging the constraints of ORCA-initialized training \cite{vandenbergReciprocalNBodyCollision2011}, we integrated custom agent behaviors that reflect real-world observations, such as (1) stop-and-go actions during hardware installation, (2) pacing in place, (3) close-range back-and-forth movements for tool retrieval, (4) volatile safety radius during panel lifting, etc. The model, initially pretrained on generic setups, underwent further alignment within these specialized scenarios, enhancing its relevance and application to actual site conditions. Alongside the customized scenarios, our training curriculum was enriched with factors such as a limited FoV, increased collision penalties, and increased random noise, further aligning the model's training with real-world complexities and safety considerations.

To mitigate drastic policy shifts and erratic Sim2Real behavior during curriculum training, we adapted the reward function $r_{\text{ori}}$ in \cite{chenCrowdrobotInteractionCrowdaware2019} by introducing a shift penalization. Actions $\{\mathbf{a}_1, \cdots, \mathbf{a}_m \}$ sampled within the refined action space $\mathcal{A}_f$ uses their estimated values as a score for ``preferability". Using the Plackett-Luce model (Eq. \ref{eq:3}), we convert value rankings into a probabilistic distribution (Eq. \ref{eq:4}), estimating and penalizing policy shifts via KL-divergence to ensure gradual and stable model adaptation (Eq. \ref{eq:5}).

\begin{align}
    P(\mathbf{a}_i | V ) &= \frac{V(\mathbf{a}_i)}{\Sigma_{j=1}^m V(\mathbf{a}_j)} \label{eq:3}\\
    D_{\text{KL}}(\pi_{\text{old}}\| \pi_{\text{new}}) &= \Sigma_{i=1}^m P_{\text{old}}(\mathbf{a}_i) \log \left( \frac{P_{\text{old}}(\mathbf{a}_i)}{P_{\text{new}}(\mathbf{a}_i)}  \right) \label{eq:4} \\
    r &= r_{\text{ori}} - \lambda D_{\text{KL}} \label{eq:5}
\end{align}

\section{Experiments and Results}

To physically examine the robot's performance in offering tangible support, we conducted the real-world evaluation and demonstration of the robot both qualitatively and quantitatively (1) on an actual construction site with workers and (2) in the lab sandbox. In the on-site demonstration and evaluation phase, we transported the robot to an active construction site and tested the robot’s functions through direct interaction with the workers. In the lab evaluation, we drew critical insights from the on-site demonstration and conducted a detailed quantitative analysis regarding several metrics.

\subsection{On-site demonstration and evaluation}

\begin{figure}[!htbp]
  \centering
  \includegraphics[width=\linewidth]{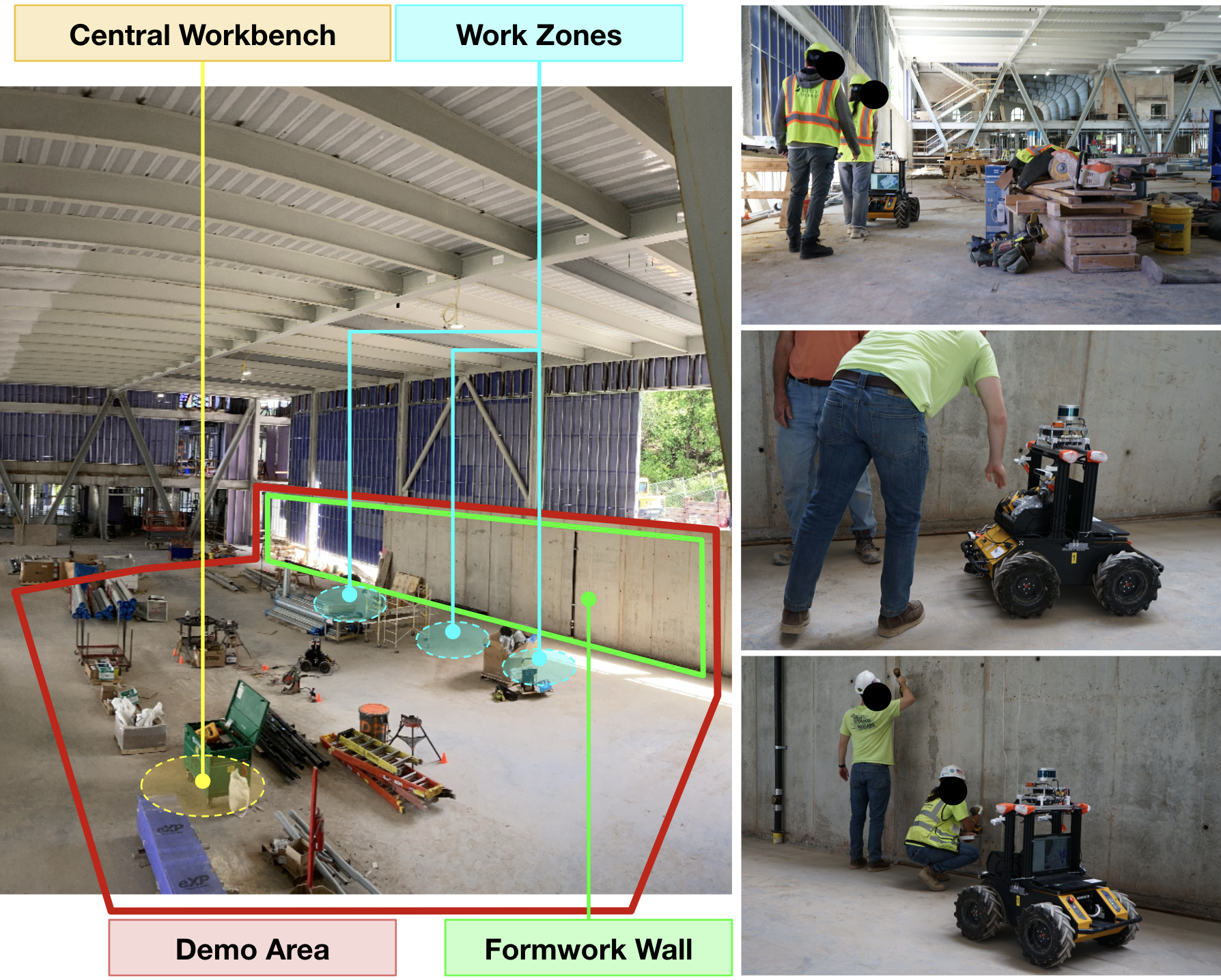}
  \caption{Demo and evaluation on an actual construction site}
  \vspace*{-3mm}
  \label{fig:site-demo}
\end{figure}

In our on-site demonstration and evaluation, we set up a 45m x 35m carpentry formwork demonstration area on an active construction floor (Fig. \ref{fig:site-demo}). This setup included a target formwork wall and several sequential work zones along it. About 15m from the wall and work zones, we designated a central workbench storing tools and metal hardware. Three construction workers participated in the evaluation, performing typical formwork installation tasks and interacting with the robot in three scripted scenarios. These scenarios, derived from our Problem Statement, involved: (1) directing the robot from the current work zone to the central workbench for tools/hardware retrieval, (2) remotely summoning the robot from the workbench to their work zone after loading, and (3) keeping the robot nearby to assist in transporting tools and hardware across installation zones. Each scenario was executed twice. The research team minimally intervened during these demonstrations, focusing primarily on documentation through cameras. The workers operated the robot using the custom-built keypad controller. Additionally, we introduced an extra test to challenge the robot’s navigational capabilities in sparsely mapped and unfamiliar areas.

The robot effectively met its support and companion roles in the initial three scenarios. Our team qualitatively noted occasional navigation pauses, particularly when the robot encountered passageways near its threshold width amidst construction objects. Regarding social comfort, there were no significant disruptions observed in the workers’ activities within the active zones, though we deferred to the workers for their perspectives on this aspect. The robot typically adjusted its path, maintaining a distance of one to two steps from the workers during active navigation. In cases where a team member deliberately and suddenly obstructed its path, the robot promptly halted and executed a brief detour. The robot, however, terminated early in the extra test case due to a lack of detailed mapping, and an unsuccessfully detected extruded pipes, converging to a localization error.

After the demonstration, we conducted a critical qualitative evaluation through a group interview with the participating workers. This interview covered three technical aspects of the robot, each linked to the context-specific challenges identified earlier.

\begin{itemize}
    \item \textbf{Robot navigation safety and comfort}: The questions posed included inquiries about the robot's navigation speed, the safety of working around it, and comfort levels, including any obstructions caused by its presence. Workers provided concise yet affirmative responses, generally agreeing that it was safe and comfortable to work alongside the robot. They did not perceive the robot as abrupt or intimidating and found it a seamless addition to their workflow.
    \item \textbf{Robot hardware design and usability}: Questions here revolved around the robot's hardware and system design, operational ease, and the intuitiveness of the controller. Workers acknowledged the robot's user-friendly interaction design and non-threatening appearance. They particularly appreciated our efforts to cover sharp edges with bubble foam for safety. However, there was a desire for enhanced physical capabilities, such as increased load capacity and the ability to haul carts.
    \item \textbf{Overall concept of robotic support in current workflow}: We inquired about the usefulness of the robot in carpentry formwork and other potential applications in construction. Workers validated the robot's current utility in reducing on-site travel and suggested its applicability in tasks like hardware distribution during interior decoration phases. They also expressed interest in future enhancements, such as autonomous tool/material loading features with robotic arms, suggesting potential for expanded manipulation functionalities.
\end{itemize}

These insights not only affirmed our initial proposal but also opened up intriguing avenues for future research and development in the realm of supportive and assistive work companion robots in heavy manual construction work.

\subsection{Quantitative lab evaluation}

\begin{figure}[!htbp]
  \centering
  \includegraphics[width=\linewidth]{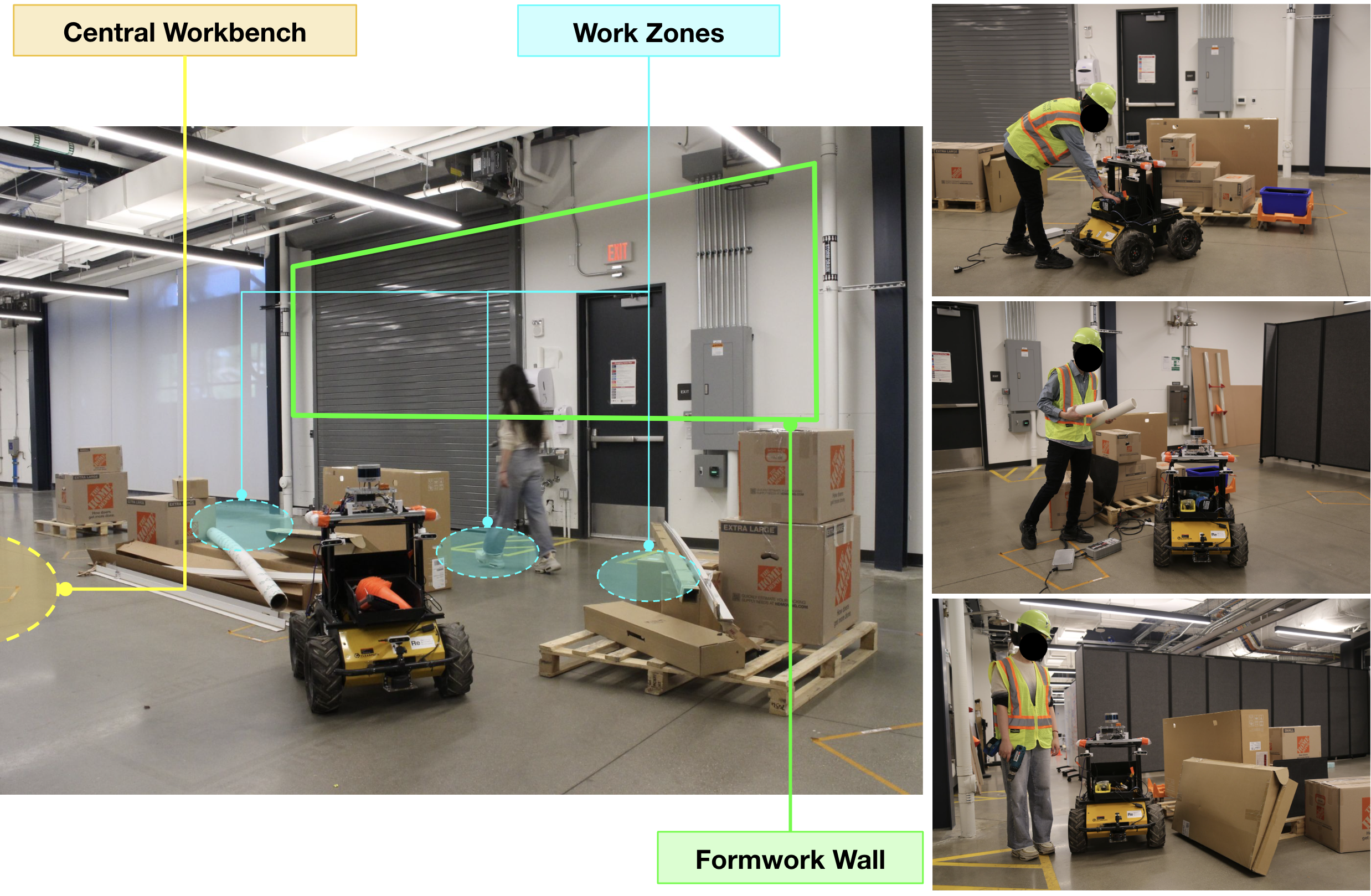}
  \caption{Lab sandbox environment evaluation}
  \vspace*{-3mm}
  \label{fig:lab-eval}
\end{figure}

\begin{figure}[!htbp]
  \centering
  \includegraphics[width=\linewidth]{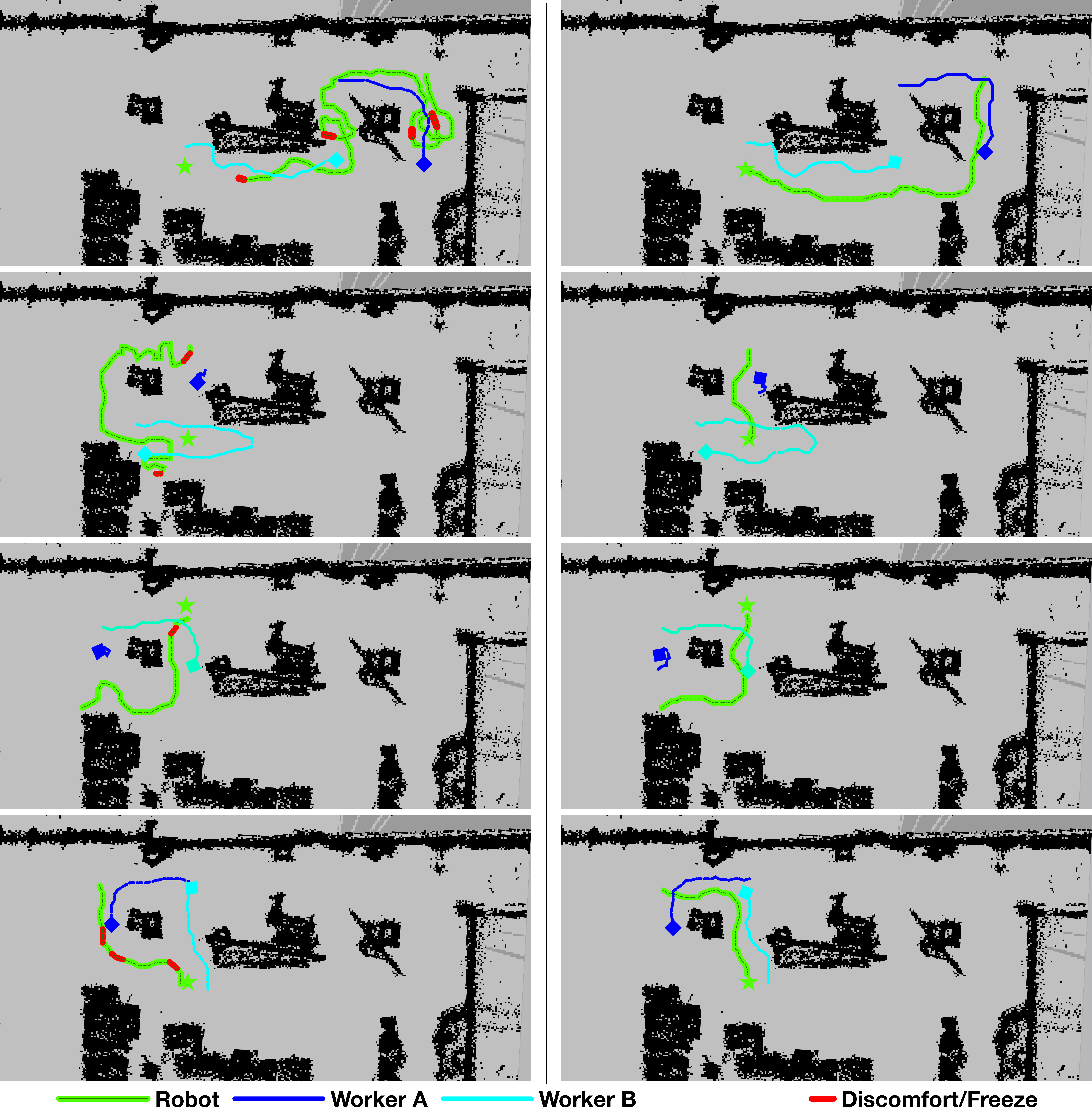}
  \caption{Trajectory comparison between RL disabled (left) and enabled (right)}
  \vspace*{-3mm}
  \label{fig:lab-eval-traj}
\end{figure}

In our laboratory, we carried out quantitative analyses and ablation studies to further validate the robot's support capabilities, particularly assessing the necessity of employing and finetuning RL in our envisioned support scenario (Fig. \ref{fig:lab-eval}). This included a performance comparison between the robot operating with and without the RL-based social navigation layer, contrasting the RL-driven hierarchical motion planning framework with basic efficiency-driven collision avoidance (Fig. \ref{fig:lab-eval-traj}).

A sandbox environment was established, maximally mirroring the on-site conditions and challenges. Evaluation metrics focused on the rate of comfortable encounters and delivery, defined by criteria such as worker proximity, movement abruptness, navigation delays, unwanted surprises, and work obstructions. Additional metrics like collision rate, success rate, freeze rate, dangerous movement rate, and travel time were also documented for a holistic performance assessment.

We designed two scenarios with 10 distinct test cases each, reflecting diverse worker activities and robot objectives. Two team members with context knowledge and safety training simulated worker behaviors—working beside walls, standing still, and moving around the site. Each case was executed six times, alternating among contextual RL, vanilla RL, and non-RL runs, in a randomized, blind test setup to minimize subjective bias in evaluating the robot's interactions. The robot's performance was carefully assessed against aforementioned criteria. Particularly, the first scenario assessed the robot's ability to navigate between the central workbench and work zones around workers amidst obstacles, while the second evaluated its capacity to accompany workers in constrained spaces between work zones.

Our empirical findings (\autoref{tab:comf}) underscore the effectiveness and necessity of the RL-based social navigation layer for enhancing the robot's support capabilities in cluttered, worker-centric environments. In scenarios not enacting RL, the comfortable encounter and delivery rates markedly decreased by approximately 0.4 and 0.15 respectively, While RL implementation did not significantly reduce major/minor collision rates (\autoref{tab:other}), it substantially decreased obstructions to workers, robot freezes and tangentially dangerous movements, resulting in smoother and more socially aware and comfortable support around workers, which validated the initial design intention.

\begin{table}[!htbp]
    \centering
    \captionsetup{format=plain,}
    \begin{tabular}{| l | c | c |}
    \hline
         & Comf. Delivery & Comf. Encounter \\
    \hline
    DWA & 0.82 & 0.56  \\
    \hline
    DWA+SARL (Vanilla) & 0.95 & 0.87 \\
    \hline
    \textbf{DWA+SARL (Contextual)} & \textbf{0.97}  & \textbf{0.91} \\
    \hline
    \end{tabular}
    \caption{Quantitative results in lab evaluation regarding comfort-related metrics.}
    
    \vspace*{-3mm}
    \label{tab:comf}
\end{table}

\begin{table}[!htbp]
    \centering
    \captionsetup{format=plain,}
    \begin{tabular}{| l | c | c | c | c |}
    \hline
         & Success & Collision & Freeze & Time \\
    \hline
    DWA & 0.83 & 0.05 & 0.95 & 29.40 \\
    \hline
    DWA+SARL (Vanilla) & 1.00 & 0.07 & \textbf{0.23} & 25.59 \\
    \hline
    \textbf{DWA+SARL (Contextual)} & 1.00 & 0.05 & \textbf{0.10} & 25.89\\
    \hline
    \end{tabular}
    
    \caption{Quantitative results in lab evaluation. ``Success”: successful delivery rate. ``Collision”: major/minor collision rate. ``Freeze": robot freeze rate. ``Time": navigation time (seconds) between two benchmarking work zones}
    \vspace*{-3mm}
    \label{tab:other}
\end{table}

Further analysis reveals that our contextualized RL model, compared to the vanilla RL model trained only in generic settings, shows an improvement in the comfortable encounter rate and a decrease in the robot freeze rate, resulting in less obstruction or interruption when offering support. This enhancement is attributed to the additional alignment tailored to work-specific scenarios and cases, which better equips the robot to handle unique patterns and attributes in our context.

\section{Conclusion and Future Work}
This study marks a considerate departure from robot-centric automation in the heavily manual construction setting, introducing a human-centric ``work companion rover" designed to assist existing human effort. Leveraging advancements in reinforcement learning, the rover is deeply contextualized in the material and social aspects of an actual construction work, in the hope of offering tangible and meaningful help to heavily manual carpentry workers. Our empirical results underscore the robot's effectiveness in real-world scenarios, offering a novel assistive prototype while highlighting a promising avenue for more diverse future research in similarly labor-intensive work contexts. Besides the current robotic support function and scenario, the developed robotic framework is possible to serve as an expandable navigation base for other supportive scenarios, approaches, and applications, calling for a future of labor ecologically combining human skill and expertise with advances in AI and robotics.

\vspace*{-5mm}

\addtolength{\textheight}{-12cm}   







\bibliographystyle{IEEEtran}
\bibliography{IROS24.bib}

\begin{thebibliography}{10}
\providecommand{\url}[1]{#1}
\csname url@rmstyle\endcsname
\providecommand{\newblock}{\relax}
\providecommand{\bibinfo}[2]{#2}
\providecommand\BIBentrySTDinterwordspacing{\spaceskip=0pt\relax}
\providecommand\BIBentryALTinterwordstretchfactor{4}
\providecommand\BIBentryALTinterwordspacing{\spaceskip=\fontdimen2\font plus
\BIBentryALTinterwordstretchfactor\fontdimen3\font minus \fontdimen4\font\relax}
\providecommand\BIBforeignlanguage[2]{{%
\expandafter\ifx\csname l@#1\endcsname\relax
\typeout{** WARNING: IEEEtran.bst: No hyphenation pattern has been}%
\typeout{** loaded for the language `#1'. Using the pattern for}%
\typeout{** the default language instead.}%
\else
\language=\csname l@#1\endcsname
\fi
#2}}

\bibitem{shehataImprovingConstructionLabor2011}
M.~E. Shehata and K.~M. {El-Gohary}, ``Towards improving construction labor productivity and projects' performance,'' \emph{Alexandria Engineering Journal}, vol.~50, no.~4, pp. 321--330, Dec. 2011.

\bibitem{fagbenroInfluencePrefabricatedConstruction2023}
R.~K. Fagbenro, R.~Y. Sunindijo, C.~Illankoon, and S.~Frimpong, ``Influence of {{Prefabricated Construction}} on the {{Mental Health}} of {{Workers}}: {{Systematic Review}},'' \emph{European Journal of Investigation in Health, Psychology and Education}, vol.~13, no.~2, pp. 345--363, Feb. 2023.

\bibitem{wuCriticalReviewUse2016}
P.~Wu, J.~Wang, and X.~Wang, ``A critical review of the use of 3-{{D}} printing in the construction industry,'' \emph{Automation in Construction}, vol.~68, pp. 21--31, Aug. 2016.

\bibitem{gharbiaRoboticTechnologiesOnsite2020}
M.~Gharbia, A.~{Chang-Richards}, Y.~Lu, R.~Y. Zhong, and H.~Li, ``Robotic technologies for on-site building construction: {{A}} systematic review,'' \emph{Journal of Building Engineering}, vol.~32, p. 101584, Nov. 2020.

\bibitem{liuBriefReviewRobotic2018}
T.~Liu, H.~Zhou, Y.~Du, J.~Zhang, J.~Zhao, and Y.~Li, ``A {{Brief Review}} on {{Robotic Floor-Tiling}},'' in \emph{{{IECON}} 2018 - 44th {{Annual Conference}} of the {{IEEE Industrial Electronics Society}}}.\hskip 1em plus 0.5em minus 0.4em\relax Washington, DC: IEEE, Oct. 2018, pp. 5583--5588.

\bibitem{asadiPictobotCooperativePainting2018}
E.~Asadi, B.~Li, and I.-M. Chen, ``Pictobot: {{A Cooperative Painting Robot}} for {{Interior Finishing}} of {{Industrial Developments}},'' \emph{IEEE Robotics \& Automation Magazine}, vol.~25, no.~2, pp. 82--94, June 2018.

\bibitem{fangArchitecturalFrameworkDistributed2020}
Z.~Fang, Y.~Wu, A.~Hassonjee, A.~Bidgoli, and D.~Cardoso~Llach, ``Towards an {{Architectural Framework}} for {{Distributed}}, {{Robotically Assisted Construction}}: {{Using Reinforcement Learning}} to {{Support Scalable Multi-Drone Construction}} in {{Dynamic Environments}}.'' in \emph{Proceedings of the 40th {{Annual Conference}} of the {{Association}} of {{Computer Aided Design}} in {{Architecture}} ({{ACADIA}})}, vol. I: Technical Papers.\hskip 1em plus 0.5em minus 0.4em\relax Online and Global: CUMINCAD, Oct. 2020, pp. 320--329.

\bibitem{bockFutureConstructionAutomation2015}
T.~Bock, ``The future of construction automation: {{Technological}} disruption and the upcoming ubiquity of robotics,'' \emph{Automation in Construction}, vol.~59, pp. 113--121, Nov. 2015.

\bibitem{suttonReinforcementLearningIntroduction2018}
R.~S. Sutton and A.~G. Barto, \emph{Reinforcement Learning: {{An}} Introduction}.\hskip 1em plus 0.5em minus 0.4em\relax MIT press, 2018.

\bibitem{mingyuemaHumanRobotTeamingConcepts2018}
L.~Mingyue~Ma, T.~Fong, M.~J. Micire, Y.~K. Kim, and K.~Feigh, ``Human-{{Robot Teaming}}: {{Concepts}} and {{Components}} for {{Design}},'' in \emph{Field and {{Service Robotics}}}, M.~Hutter and R.~Siegwart, Eds.\hskip 1em plus 0.5em minus 0.4em\relax Cham: Springer International Publishing, 2018, pp. 649--663.

\bibitem{calderitaTHERAPISTAutonomousSocially2014}
L.~V. Calderita, L.~J. Manso, P.~Bustos, C.~{Su{\'a}rez-Mej{\'i}as}, F.~Fern{\'a}ndez, and A.~Bandera, ``{{THERAPIST}}: {{Towards}} an autonomous socially interactive robot for motor and neurorehabilitation therapies for children,'' \emph{JMIR rehabilitation and assistive technologies}, vol.~1, no.~1, p.~e1, Oct. 2014.

\bibitem{fasolaSociallyAssistiveRobot2013}
J.~Fasola and M.~J. Matari{\'c}, ``A socially assistive robot exercise coach for the elderly,'' \emph{Journal of Human-Robot Interaction}, vol.~2, no.~2, pp. 3--32, June 2013.

\bibitem{iioHumanLikeGuideRobot2020}
T.~Iio, S.~Satake, T.~Kanda, K.~Hayashi, F.~Ferreri, and N.~Hagita, ``Human-{{Like Guide Robot}} that {{Proactively Explains Exhibits}},'' \emph{International Journal of Social Robotics}, vol.~12, no.~2, pp. 549--566, May 2020.

\bibitem{kruijffExperienceSystemDesign2014}
G.-J.~M. Kruijff, M.~Jan{\'i}{\v c}ek, S.~Keshavdas, B.~Larochelle, H.~Zender, N.~J. Smets, T.~Mioch, M.~A. Neerincx, J.~V. Diggelen, F.~Colas, \emph{et~al.}, ``Experience in system design for human-robot teaming in urban search and rescue,'' in \emph{Field and Service Robotics: {{Results}} of the 8th International Conference}.\hskip 1em plus 0.5em minus 0.4em\relax Springer, 2014, pp. 111--125.

\bibitem{hopkoHumanFactorsConsiderations2022}
S.~Hopko, J.~Wang, and R.~Mehta, ``Human factors considerations and metrics in shared space human-robot collaboration: {{A}} systematic review,'' \emph{Frontiers in Robotics and AI}, vol.~9, p. 799522, 2022.

\bibitem{pandeyMobileRobotNavigation2017}
A.~Pandey, S.~Pandey, and {\relax DR}.~Parhi, ``Mobile robot navigation and obstacle avoidance techniques: {{A}} review,'' \emph{Int Rob Auto J}, vol.~2, no.~3, p. 00022, 2017.

\bibitem{mavrogiannisCoreChallengesSocial2021}
C.~Mavrogiannis, F.~Baldini, A.~Wang, D.~Zhao, P.~Trautman, A.~Steinfeld, and J.~Oh, ``Core {{Challenges}} of {{Social Robot Navigation}}: {{A Survey}},'' Mar. 2021.

\bibitem{chenAdoptionSelfdrivingDelivery2021}
C.~Chen, E.~Demir, Y.~Huang, and R.~Qiu, ``The adoption of self-driving delivery robots in last mile logistics,'' \emph{Transportation Research Part E: Logistics and Transportation Review}, vol. 146, p. 102214, Feb. 2021.

\bibitem{williamssonBusinessModelDesign2022}
J.~Williamsson, ``Business model design for campus-based autonomous deliveries -- {{A Swedish}} case study,'' \emph{Research in Transportation Business \& Management}, vol.~43, p. 100758, June 2022.

\bibitem{wetzelUseBostonDynamics2022}
{\relax EM}.~Wetzel, J.~Liu, T.~Leathem, and A.~Sattineni, ``The use of boston dynamics {{SPOT}} in support of {{LiDAR}} scanning on active construction sites,'' in \emph{{{ISARC}}. {{Proceedings}} of the International Symposium on Automation and Robotics in Construction}, vol.~39.\hskip 1em plus 0.5em minus 0.4em\relax IAARC Publications, 2022, pp. 86--92.

\bibitem{singamaneniSurveySociallyAware2024}
P.~T. Singamaneni, P.~{Bachiller-Burgos}, L.~J. Manso, A.~Garrell, A.~Sanfeliu, A.~Spalanzani, and R.~Alami, ``A survey on socially aware robot navigation: {{Taxonomy}} and future challenges,'' \emph{The International Journal of Robotics Research}, p. 02783649241230562, Feb. 2024.

\bibitem{everettMotionPlanningDynamic2018}
M.~Everett, Y.~F. Chen, and J.~P. How, ``Motion planning among dynamic, decision-making agents with deep reinforcement learning,'' in \emph{{{IEEE}}/{{RSJ}} International Conference on Intelligent Robots and Systems ({{IROS}})}, Madrid, Spain, Sept. 2018.

\bibitem{chenCrowdrobotInteractionCrowdaware2019}
C.~Chen, Y.~Liu, S.~Kreiss, and A.~Alahi, ``Crowd-robot interaction: {{Crowd-aware}} robot navigation with attention-based deep reinforcement learning,'' in \emph{2019 International Conference on Robotics and Automation ({{ICRA}})}.\hskip 1em plus 0.5em minus 0.4em\relax IEEE, 2019, pp. 6015--6022.

\bibitem{liuDecentralizedStructuralRNNRobot2021}
S.~Liu, P.~Chang, W.~Liang, N.~Chakraborty, and K.~{Driggs-Campbell}, ``Decentralized {{Structural-RNN}} for {{Robot Crowd Navigation}} with {{Deep Reinforcement Learning}},'' in \emph{2021 {{IEEE International Conference}} on {{Robotics}} and {{Automation}} ({{ICRA}})}.\hskip 1em plus 0.5em minus 0.4em\relax Xi'an, China: IEEE, May 2021, pp. 3517--3524.

\bibitem{stolerT2FPVDatasetMethod2023}
B.~Stoler, M.~Jana, S.~Hwang, and J.~Oh, ``{{T2FPV}}: {{Dataset}} and method for correcting first-person view errors in pedestrian trajectory prediction,'' in \emph{2023 {{IEEE}}/{{RSJ}} International Conference on Intelligent Robots and Systems ({{IROS}})}.\hskip 1em plus 0.5em minus 0.4em\relax IEEE, 2023, pp. 4037--4044.

\bibitem{zhangLOAMLidarOdometry2014}
J.~Zhang and S.~Singh, ``{{LOAM}}: {{Lidar Odometry}} and {{Mapping}} in {{Real-time}},'' in \emph{Robotics: {{Science}} and {{Systems X}}}.\hskip 1em plus 0.5em minus 0.4em\relax {Robotics: Science and Systems Foundation}, July 2014.

\bibitem{shanLIOSAMTightlycoupledLidar2020}
T.~Shan, B.~Englot, D.~Meyers, W.~Wang, C.~Ratti, and D.~Rus, ``{{LIO-SAM}}: {{Tightly-coupled Lidar Inertial Odometry}} via {{Smoothing}} and {{Mapping}},'' in \emph{2020 {{IEEE}}/{{RSJ International Conference}} on {{Intelligent Robots}} and {{Systems}} ({{IROS}})}, Oct. 2020, pp. 5135--5142.

\bibitem{everettRobotDesignedSocially2017}
M.~F. Everett, ``Robot designed for socially acceptable navigation,'' Ph.D. dissertation, Massachusetts Institute of Technology, 2017.

\bibitem{wangYOLOv7TrainableBagoffreebies2022}
C.-Y. Wang, A.~Bochkovskiy, and H.-Y.~M. Liao, ``{{YOLOv7}}: {{Trainable}} bag-of-freebies sets new state-of-the-art for real-time object detectors,'' \emph{arXiv preprint arXiv:2207.02696}, 2022.

\bibitem{schubertDBSCANRevisitedRevisited2017}
E.~Schubert, J.~Sander, M.~Ester, H.~P. Kriegel, and X.~Xu, ``{{DBSCAN}} revisited, revisited: Why and how you should (still) use {{DBSCAN}},'' \emph{ACM Transactions on Database Systems (TODS)}, vol.~42, no.~3, pp. 1--21, 2017.

\bibitem{wojkeSimpleOnlineRealtime2017}
N.~Wojke, A.~Bewley, and D.~Paulus, ``Simple online and realtime tracking with a deep association metric,'' in \emph{2017 {{IEEE}} International Conference on Image Processing ({{ICIP}})}.\hskip 1em plus 0.5em minus 0.4em\relax IEEE, 2017, pp. 3645--3649.

\bibitem{foxDynamicWindowApproach1997}
D.~Fox, W.~Burgard, and S.~Thrun, ``The dynamic window approach to collision avoidance,'' \emph{IEEE Robotics \& Automation Magazine}, vol.~4, no.~1, pp. 23--33, 1997.

\bibitem{chenSociallyAwareMotion2017}
Y.~F. Chen, M.~Everett, M.~Liu, and J.~P. How, ``Socially aware motion planning with deep reinforcement learning,'' in \emph{2017 {{IEEE}}/{{RSJ International Conference}} on {{Intelligent Robots}} and {{Systems}} ({{IROS}})}, Sept. 2017, pp. 1343--1350.

\bibitem{vandenbergReciprocalNBodyCollision2011}
J.~{van den Berg}, S.~J. Guy, M.~Lin, and D.~Manocha, ``Reciprocal n-{{Body Collision Avoidance}},'' in \emph{Robotics {{Research}}}, ser. Springer {{Tracts}} in {{Advanced Robotics}}, C.~Pradalier, R.~Siegwart, and G.~Hirzinger, Eds.\hskip 1em plus 0.5em minus 0.4em\relax Berlin, Heidelberg: Springer, 2011, pp. 3--19.

\end{thebibliography}

\end{document}